\documentclass[runningheads]{llncs}
\usepackage{graphicx}
\usepackage{comment}
\usepackage{amsmath,amssymb} 
\usepackage{color}
\usepackage[colorlinks=true]{hyperref}

\graphicspath{ {./images/} }
\graphicspath{ {./supp/} }\usepackage{multirow, subcaption, gensymb}
\usepackage[font=small,labelfont=bf]{caption}
\begin{document}
\pagestyle{headings}
\mainmatter
\def\ECCVSubNumber{2986}  

\title{Personalized Face Modeling for Improved Face \\ Reconstruction and Motion Retargeting} 

\titlerunning{Personalized Face Modeling for Reconstruction and Retargeting}
%
\author{Bindita Chaudhuri\inst{1}\thanks{This work was done when the author visited Microsoft.} \and
Noranart Vesdapunt\inst{2} \and \\
Linda Shapiro\inst{1} \and 
Baoyuan Wang\inst{2}
}
\authorrunning{B. Chaudhuri et al.}
\institute{University of Washington\\
\email{\{bindita,shapiro\}@cs.washington.edu}\and
Microsoft Cloud and AI\\
\email{\{noves,baoyuanw\}@microsoft.com}}
\maketitle

\begin{abstract}
Traditional methods for image-based 3D face reconstruction and facial motion retargeting fit a 3D morphable model (3DMM) to the face, which has limited modeling capacity and fail to generalize well to in-the-wild data. Use of deformation transfer or multilinear tensor as a personalized 3DMM for blendshape interpolation does not address the fact that facial expressions result in different local and global skin deformations in different persons. Moreover, existing methods learn a single albedo per user which is not enough to capture the expression-specific skin reflectance variations. We propose an end-to-end framework that jointly learns a personalized face model per user and per-frame facial motion parameters from a large corpus of in-the-wild videos of user expressions. Specifically, we learn user-specific expression blendshapes and dynamic (expression-specific) albedo maps by predicting personalized corrections on top of a 3DMM prior. We introduce novel training constraints to ensure that the corrected blendshapes retain their semantic meanings and the reconstructed geometry is disentangled from the albedo. Experimental results show that our personalization accurately captures fine-grained facial dynamics in a wide range of conditions and efficiently decouples the learned face model from facial motion, resulting in more accurate face reconstruction and facial motion retargeting compared to state-of-the-art methods.
\end{abstract}

\section{Introduction}
With the ubiquity of mobile phones, AR/VR headsets and video games, communication through facial gestures has become very popular, leading to extensive research in problems like 2D face alignment, 3D face reconstruction and facial motion retargeting. A major component of these problems is to estimate the 3D face, i.e., face geometry, appearance, expression, head pose and scene lighting, from 2D images or videos. 3D face reconstruction from monocular images is ill-posed by nature, so a typical solution is to leverage a parametric 3D morphable model (3DMM) trained on a limited number of 3D face scans as prior knowledge \cite{3DMM,BFM,Vlasic:2005:FTM,FLAME,3DShape15,3Dfacerig16,face2face,deng2019accurate,inversefacenet}. However, the low dimensional space limits their modeling capacity as shown in \cite{selfsupervised_ayush,nonlinear3dmm2018,Jackson_2017_ICCV} and scalability using more 3D scans is expensive. Similarly, the texture model of a generic 3DMM is learned in a controlled environment and does not generalize well to in-the-wild images. Tran et al. \cite{nonlinear3dmm2018,nonlinear3dmm2019} overcomes these limitations by learning a non-linear 3DMM from a large corpus of in-the-wild images. Nevertheless, these reconstruction-based approaches do not easily support facial motion retargeting.

In order to perform tracking for retargeting, blendshape interpolation technique is usually adopted where the users' blendshapes are obtained by deformation transfer \cite{deformationtransfer}, but this alone cannot reconstruct expressions realistically as shown in \cite{3Dfacerig16,examplebasedrigging}. Another popular technique is to use a multilinear tensor-based 3DMM \cite{Vlasic:2005:FTM,DDEregression,Cao:2018:SRF}, where the expression is coupled with the identity implying that same identities should share the same expression blendshapes. However, we argue that facial expressions are characterized by different skin deformations on different persons due to difference in face shape, muscle movements, age and other factors. This kind of user-specific local skin deformations cannot be accurately represented by a linear combination of predefined blendshapes. For example, smiling and raising eyebrows create different cheek folds and forehead wrinkle patterns respectively on different persons, which cannot be represented by simple blendshape interpolation and require correcting the corresponding blendshapes. Some optimization-based approaches \cite{examplebasedrigging,3Dfacerig16,bouaziz_avatarfromvideo,mocap} have shown that modeling user-specific blendshapes indeed results in a significant improvement in the quality of face reconstruction and tracking. However, these approaches are computationally slow and require additional preprocessing (e.g. landmark detection) during test time, which significantly limits real-time applications with in-the-wild data on the edge devices. The work \cite{MultiFaceRetarget19} trains a deep neural network instead to perform retargeting in real-time on typical mobile phones, but its use of predefined 3DMM limits its face modeling accuracy. Tewari et al. \cite{fml} leverage in-the-wild videos to learn face identity and appearance models from scratch, but they still use expression blendshapes generated by deformation transfer.

Moreover, existing methods learn a single albedo map for a user. The authors in \cite{disney} have shown that skin reflectance changes with skin deformations, but it is not feasible to generate a separate albedo map for every expression during retargeting. Hence it is necessary to learn the static reflectance separately, and associate the expression-specific dynamic reflectance with the blendshapes so that the final reflectance can be obtained by interpolation similar to blendshape interpolation, as in \cite{pagan}. Learning dynamic albedo maps in addition to static albedo map also helps to capture the fine-grained facial expression details like folds and wrinkles \cite{pagan_previous}, thereby resulting in reconstruction of higher fidelity. 

To address these issues, we introduce a novel end-to-end framework that leverages a large corpus of in-the-wild user videos to jointly learn personalized face modeling and face tracking parameters. Specifically, we design a modeling network which learns geometry and reflectance corrections on top of a 3DMM prior to generate user-specific expression blendshapes and dynamic (expression-specific) albedo maps. In order to ensure proper disentangling of the geometry from the albedo, we introduce the face parsing loss inspired by \cite{Zhu_2020_CVPR}. Note that \cite{Zhu_2020_CVPR} uses parsing loss in a fitting based framework whereas we use it in a learning based framework. We also ensure that the corrected blendshapes retain their semantic meanings by restricting the corrections to local regions using attention maps and by enforcing a blendshape gradient loss. We design a separate tracking network which predicts the expression blendshape coefficients, head pose and scene lighting parameters. The decoupling between the modeling and tracking networks enables our framework to perform reconstruction as well as retargeting (by tracking one user and transferring the facial motion to another user's model). Our main contributions are:

\begin{enumerate}
\itemsep0em 
    \item We propose a deep learning framework to learn user-specific expression blendshapes and dynamic albedo maps that accurately capture the complex user-specific expression dynamics and high-frequency details like folds and wrinkles, thereby resulting in photorealistic 3D face reconstruction.
    \item We bring two novel constraints into the end-to-end training: face parsing loss to reduce the ambiguity between geometry and reflectance and blendshape gradient loss to retain the semantic meanings of the corrected blendshapes. 
    \item Our framework jointly learns user-specific face model and user-independent facial motion in disentangled form, thereby supporting motion retargeting.
\end{enumerate}

\section{Related Work}

\noindent \textbf{Face Modeling:}
Methods like \cite{incrementalfacetracking,mesoscopicgeometry,deep_appearance_model,RingNet:CVPR:2019,realtime:cnn:animation,Liu_2018_CVPR,tran2017regressing} leverage user images captured with varying parameters (e.g. multiple viewpoints, expressions etc.) at least during training with the aim of user-specific 3D face reconstruction (not necessarily retargeting). Monocular video-based optimization techniques for 3D face reconstruction \cite{pablo_geomfromvideo,3Dfacerig16} leverage the multi-frame consistency to learn the facial details. For single image based reconstruction, traditional methods \cite{sota_facereconstruction} regress the parameters of a 3DMM and then learn corrective displacement \cite{mesoscopicgeometry,3DFaceRecTIP18,cnnreconst} or normal maps \cite{wildphotobasedreconstruction,Richardson} to capture the missing details. Recently, several deep learning based approaches have attempted to overcome the limited representation power of 3DMM. Tran et al. \cite{nonlinear3dmm2018,nonlinear3dmm2019} proposed to train a deep neural network as a non-linear 3DMM. Tewari et al. \cite{selfsupervised_ayush} proposed to learn shape and reflectance correctives on top of the linear 3DMM. In \cite{fml}, Tewari et al. learn new identity and appearance models from videos. However, these methods use expression blendshapes obtained by deformation transfer \cite{deformationtransfer} from a generic 3DMM to their own face model and do not optimize the blendshapes based on the user's identity. In addition, these methods predict a single static albedo map to represent the face texture, which fail to capture adequate facial details.

\noindent \textbf{Personalization:}
Optimization based methods like \cite{examplebasedrigging,bouaziz_avatarfromvideo,chencao_imagebasedavatar} have demonstrated the need to optimize the expression blendshapes based on user-specific facial dynamics. These methods alternately update the blendshapes and the corresponding coefficients to accurately fit some example poses in the form of 3D scans or 2D images. For facial appearance, existing methods either use a generic texture model with linear or learned bases or use a GAN \cite{GANFit} to generate a static texture map. But different expressions result in different texture variations, and Nagano et al. \cite{pagan} and Olszewski et al. \cite{pagan_previous} addressed this issue by using a GAN to predict the expression-specific texture maps given the texture map in neutral pose. However, the texture variations with expression also vary from person to person. Hence, hallucinating an expression-specific texture map for a person by learning the expression dynamics of other persons is not ideal. Besides, these methods requires fitted geometry as a preprocessing step, thereby limiting the accuracy of the method by the accuracy of the geometry fitting mechanism.

\noindent \textbf{Face Tracking and Retargeting:}
Traditional face tracking and retargeting methods \cite{bouaziz_realtimefaceanimation,onlinemodelingBouaziz,ontheflycorrectives} generally optimize the face model parameters with occasional correction of the expression blendshapes using depth scans. Recent deep learning based tracking frameworks like \cite{face2face,incrementalfacetracking,MultiFaceRetarget19,deepvideopotrait} either use a generic face model and fix the model during tracking, or alternate between tracking and modeling until convergence. We propose to perform joint face modeling and tracking with novel constraints to disambiguate the tracking parameters from the model.

\begin{figure}[t]
    \centering
    \includegraphics[width=1.0\textwidth]{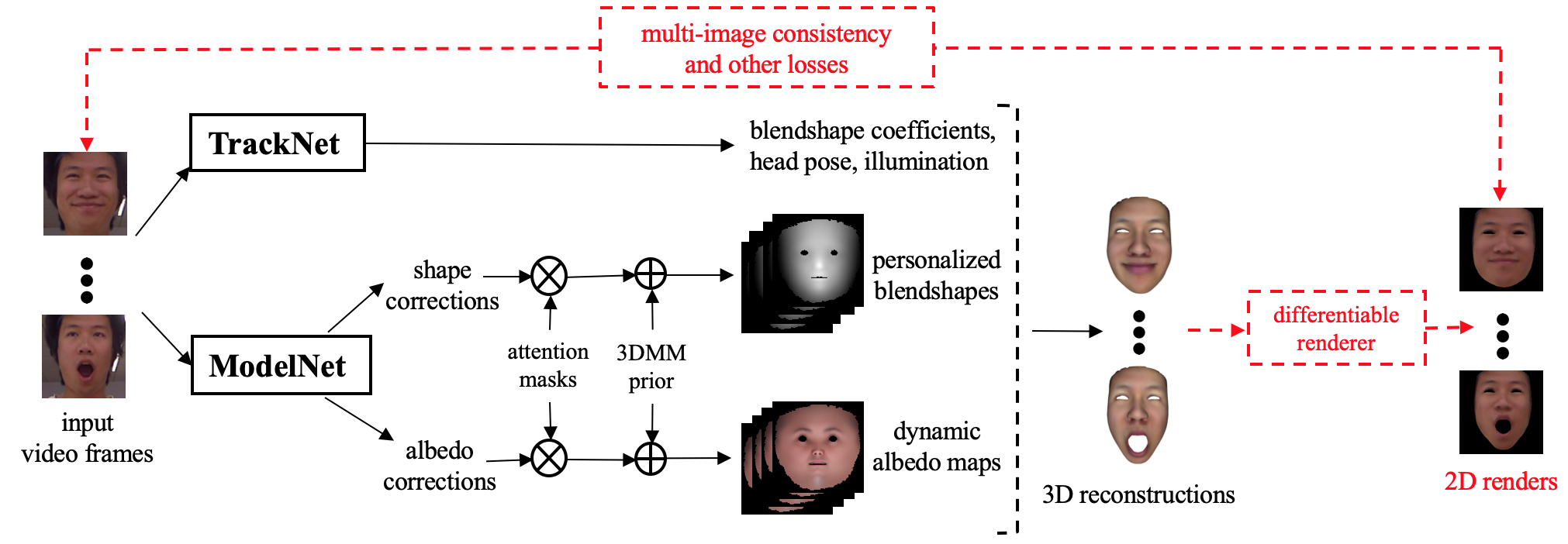}
    \caption{\textbf{Our end-to-end framework}. Our framework takes frames from in-the-wild video(s) of a user as input and generates per-frame tracking parameters via the \textit{TrackNet} and personalized face model via the \textit{ModelNet}. The networks are trained together in an end-to-end manner (marked in red) by projecting the reconstructed 3D outputs into 2D using a differentiable renderer and computing multi-image consistency losses and other regularization losses.}
    \label{fig:pipeline}
\end{figure}

\section{Methodology}
\subsection{Overview}
Our network architecture, as shown in Fig. \ref{fig:pipeline}, has two parts: 1) \textit{ModelNet} which learns to capture the user-specific facial details and 2) \textit{TrackNet} which learns to capture the user-independent facial motion. The networks are trained together in an end-to-end manner using multi-frame images of different identities, i.e., multiple images $\{I_1, \ldots, I_N\}$ of the same person sampled from a video in each mini-batch. We leverage the fact that the person's facial geometry and appearance remain unchanged across all the frames in a video, whereas the facial expression, head pose and scene illumination change on a per-frame basis. The \textit{ModelNet} extracts a common feature from all the $N$ images to learn a user-specific face shape, expression blendshapes and dynamic albedo maps (Section \ref{sec:facemodellearning}). The \textit{TrackNet} processes each of the $N$ images individually to learn the image-specific expression blendshape coefficients, pose and illumination parameters (Section \ref{sec:jointtraining}). The predictions of \textit{ModelNet} and \textit{TrackNet} are combined to reconstruct the 3D faces and then projected to the 2D space using a differentiable renderer in order to train the network in a self-supervised manner using multi-image photometric consistency, landmark alignment and other constraints. During testing, the default settings can perform 3D face reconstruction. However, our network architecture and training strategy allow simultaneous tracking of one person's face using \textit{TrackNet} and modeling another person's face using \textit{ModelNet}, and then retarget the tracked person's facial motion to the modeled person or an external face model having similar topology as our face model.

\subsection{Learning Personalized Face Model}
\label{sec:facemodellearning}
Our template 3D face consists of a mean (neutral) face mesh $S_0$ having 12K vertices, per-vertex colors (converted to 2D mean albelo map $R_0$ using UV coordinates) and 56 expression blendshapes $\{S_1, \ldots, S_{56}\}$. Given a set of expression coefficients $\{w_1, \ldots, w_{56}\}$, the template face shape can be written as $\bar{S} = w_0 S_0 + \sum_{i=1}^{56} w_i S_i$ where $w_0 = (1 - \sum_{i=1}^{56} w_i)$. Firstly, we propose to learn an identity-specific corrective deformation $\Delta^S_0$ from the identity of the input images to convert $\bar{S}$ to identity-specific shape. Then, in order to better fit the facial expression of the input images, we learn corrective deformations $\Delta^S_i$ for each of the template blendshapes $S_i$ to get identity-specific blendshapes. Similarly, we learn a corrective albedo map $\Delta^R_0$ to convert $R_0$ to identity-specific static albedo map. In addition, we also learn corrective albedo maps $\Delta^R_i$ corresponding to each $S_i$ to get identity-specific dynamic (expression-specific) albedo maps.

In our \textit{ModelNet}, we use a shared convolutional encoder $E^{\text{model}}$ to extract features $F^{\text{model}}_n$ from each image $I_n \in \{I_1, \ldots, I_N\}$ in a mini-batch. Since all the $N$ images belong to the same person, we take an average over all the $F^{\text{model}}_n$ features to get a common feature $F^{\text{model}}$ for that person. Then, we pass $F^{\text{model}}$ through two separate convolutional decoders, $D_S^{\text{model}}$ to estimate the shape corrections $\Delta^S_0$ and $\Delta^S_i$, and $D_R^{\text{model}}$ to estimate the albedo corrections $\Delta^R_0$ and $\Delta^R_i$. We learn the corrections in the UV space instead of the vertex space to reduce the number of network parameters and preserve the contextual information. 

\noindent \textbf{User-specific expression blendshapes}
A naive approach to learn corrections on top of template blendshapes based on the user's identity would be to predict corrective values for all the vertices and add them to the template blendshapes. However, since blendshape deformation is local, we want to restrict the corrected deformation to a similar local region as the template deformation. To do this, we first apply an attention mask over the per-vertex corrections and then add it to the template blendshape. We compute the attention mask $A_i$ corresponding to the blendshape $S_i$ by calculating the per-vertex euclidean distances between $S_i$ and $S_0$, thresholding them at 0.001, normalizing them by the maximum distance, and then converting them into the UV space. We also smooth the mask discontinuities using a small amount of Gaussian blur following \cite{pagan}. Finally, we multiply $A_i$ with $\Delta^S_i$ and add it to $S_i$ to obtain a corrected $S_i$. Note that the masks are precomputed and then fixed during network operations. The final face shape is thus given by:
\begin{equation}
    S =  w_0 S_0 + \mathcal{F}(\Delta^S_0) + \sum_{i=1}^{56} w_i [S_i + \mathcal{F}(A_i \Delta^S_i)] 
\label{eq:shape}
\end{equation}
where $\mathcal{F}(\cdot)$ is a sampling function for UV space to vertex space conversion.

\noindent \textbf{User-specific dynamic albedo maps}
We use one static albedo map to represent the identity-specific neutral face appearance and 56 dynamic albedo maps, one for each expression blendshape, to represent the expression-specific face appearance. Similar to blendshape corrections, we predict 56 albedo correction maps in the UV space and add them to the static albedo map after multiplying the dynamic correction maps with the corresponding UV attention masks. Our final face albedo is thus given by:
\begin{equation}
    R = R^t_0 + \Delta^R_0 + \sum_{i=1}^{56} w_i [A_i \Delta^R_i] 
\label{eq:reflectance}
\end{equation}
where $R^t_0$ is the trainable mean albedo initialized with the mean albedo $R_0$ from our template face similar to \cite{fml}.

\subsection{Joint Modeling and Tracking}
\label{sec:jointtraining}
The \textit{TrackNet} consists of a convolutional encoder $E^{\text{track}}$ followed by multiple fully connected layers to regress the tracking parameters $\mathbf{p}_n = (\mathbf{w}_n, \mathbf{R}_n, \mathbf{t}_n, \gamma_n)$ for each image $I_n$. The encoder and fully connected layers are shared over all the $N$ images in a mini-batch. Here $\mathbf{w}_n = (w^n_0, \ldots, w^n_{56})$ is the expression coefficient vector and $\mathbf{R}_n \in SO(3)$ and $\mathbf{t}_n \in \mathbb{R}^3$ are the head rotation (in terms of Euler angles) and 3D translation respectively. $\gamma_n \in \mathbb{R}^{27}$ are the 27 Spherical Harmonics coefficients (9 per color channel) following the illumination model of \cite{fml}. 

\noindent \textbf{Training Phase}:
We first obtain a face shape $S_n$ and albedo $R_n$ for each $I_n$ by combining $S$ (equation \ref{eq:shape}) and $R$ (equation \ref{eq:reflectance}) from the \textit{ModelNet} and the expression coefficient vector $\mathbf{w}_n$ from the \textit{TrackNet}. Then, similar to \cite{GANFit,fml}, we transform the shape using head pose as $\tilde{S}_n = \mathbf{R}_n S_n + \mathbf{t}_n$ and project it onto the 2D camera space using a perspective camera model $\Phi: \mathbb{R}^3 \rightarrow \mathbb{R}^2$. Finally, we use a differentiable renderer $\mathcal{R}$ to obtain the reconstructed 2D image as $\hat{I}_n = \mathcal{R}(\tilde{S}_n, \mathbf{n}_n, R_n, \gamma_n)$ where $\mathbf{n}_n$ are the per-vertex normals. We also mark 68 facial landmarks on our template mesh which we can project onto the 2D space using $\Phi$ to compare with the ground truth 2D landmarks. 

\noindent \textbf{Testing Phase}:
The \textit{ModelNet} can take a variable number of input images of a person (due to our feature averaging technique) to predict a personalized face model. The \textit{TrackNet} executes independently on one or more images of the same person given as input to \textit{ModelNet} or a different person. For face reconstruction, we feed images of the same person to both the networks and combine their outputs as in the training phase to get the 3D faces. In order to perform facial motion retargeting, we first obtain the personalized face model of the target subject using \textit{ModelNet}. We then predict the facial motion of the source subject on a per-frame basis using the \textit{TrackNet} and combine it with the target face model. It is important to note that the target face model can be any external face model with semantically similar expression blendshapes.
 
\subsection{Loss Functions}
We train both the \textit{TrackNet} and the \textit{ModelNet} together in an end-to-end manner using the following loss function:
\begin{equation}
    L = \lambda_{\text{ph}} L_{\text{ph}} + \lambda_{\text{lm}} L_{\text{lm}} + \lambda_{\text{pa}} L_{\text{pa}} + \lambda_{\text{sd}} L_{\text{sd}} + \lambda_{\text{bg}} L_{\text{bg}} + \lambda_{\text{reg}} L_{\text{reg}}
\label{eq:loss}
\end{equation}
where the loss weights $\lambda_*$ are chosen empirically and their values are given in the supplementary material\footnote{\url{https://homes.cs.washington.edu/~bindita/personalizedfacemodeling.html}}.

\noindent \textbf{Photometric and Landmark Losses:} We use the $l_{2,1}$ \cite{nonlinear3dmm2019} loss to compute the multi-image photometric consistency loss between the input images $I_n$ and the reconstructed images $\hat{I}_n$. The loss is given by
\begin{equation}
    L_{\text{ph}} = \sum_{n=1}^N \frac{\sum_{q=1}^Q ||M_n(q) * [I_n(q) - \hat{I}_n(q)]||_2}{\sum_{q=1}^Q M_n(q)}  
\end{equation}
where $M_n$ is the mask generated by the differentiable renderer (to exclude the background, eyeballs and mouth interior) and $q$ ranges over all the pixels $Q$ in the image. In order to further improve the quality of the predicted albedo by preserving high-frequency details, we add the image (spatial) gradient loss having the same expression as the photometric loss with the images replaced by their gradients. Adding other losses as in \cite{GANFit} resulted in no significant improvement. The landmark alignment loss $L_{\text{lm}}$ is computed as the $l_2$ loss between the ground truth and predicted 68 2D facial landmarks.

\noindent \textbf{Face Parsing Loss:} The photometric and landmark loss constraints are not strong enough to overcome the ambiguity between shape and albedo in the 2D projection of a 3D face. Besides, the landmarks are sparse and often unreliable especially for extreme poses and expressions which are difficult to model because of depth ambiguity. So, we introduce the face parsing loss given by:
\begin{equation}
    L_{\text{pa}} = \sum_{n=1}^N ||I^{\text{pa}}_n - \hat{I}^{\text{pa}}_n||_2  
\end{equation}
where $I^{\text{pa}}_n$ is the ground truth parsing map generated using the method in \cite{faceparsing} and $\hat{I}^{\text{pa}}_n$ is the predicted parsing map generated as $\mathcal{R}(\tilde{S}_n, \mathbf{n}_n, T)$ with a fixed precomputed UV parsing map $T$.

\noindent \textbf{Shape Deformation Smoothness Loss:}
We employ Laplacian smoothness on the identity-specific corrective deformation to ensure that our predicted shape is locally smooth. The loss is given as:
\begin{equation}
L_{\text{sd}} = \sum_{v=1}^V \sum_{u\in \mathcal{N}_v} ||\Delta^S_0(v) - \Delta^S_0(u)||^2_2
\end{equation}
where $V$ is the total number of vertices in our mesh and $\mathcal{N}_v$ is the set of neighboring vertices directly connected to vertex $v$.

\noindent \textbf{Blendshape Gradient Loss:} Adding free-form deformation to a blendshape, even after restricting it to a local region using attention masks, can change the semantic meaning of the blendshape. However, in order to retarget tracked facial motion of one person to the blendshapes of another person, the blendshapes of both the persons should have semantic correspondence. We introduce a novel blendshape gradient loss to ensure that the deformation gradients of the corrected blendshapes are similar to those of the template blendshapes. The loss is given by:
\begin{equation}
    L_{\text{bg}} = \sum_{i=1}^{56} ||\mathbf{G}_{S_0 \rightarrow (S_i+\Delta^S_i)} - \mathbf{G}_{S_0 \rightarrow S_i} ||^2_2
\end{equation}
where $\mathbf{G}_{a \rightarrow b}$ denotes the gradient of the deformed mesh $b$ with respect to original mesh $a$. Details about how to compute $\mathbf{G}$ can be found in \cite{examplebasedrigging}.

\noindent \textbf{Tracking Parameter Regularization Loss:} We use sigmoid activation at the output of the expression coefficients and regularize the coefficients using $l_1$ loss ($L^{w}_{\text{reg}}$) to ensure sparse coefficients in the range $[0,1]$. In order to disentangle the albedo from the lighting, we use a lighting regularization loss given by:
\begin{equation}
    L^{\gamma}_{\text{reg}} = ||\gamma - \gamma_{\text{mean}}||_2 + \lambda_\gamma ||\gamma||_2
\end{equation}
where the first term ensures that the predicted light is mostly monochromatic and the second term restricts the overall lighting. We found that regularizing the illumination automatically resulted in albedo consistency, so we don't use any additional albedo loss. Finally, $L_{\text{reg}} = L^{w}_{\text{reg}} + L^{\gamma}_{\text{reg}}$.

\section{Experimental Setup}

\noindent \textbf{Datasets:}
We train our network using two datasets: 1) VoxCeleb2 \cite{voxceleb2} and 2) ExpressiveFaces. We set aside 10\% of each dataset for testing. VoxCeleb2 has more than 140k videos of about 6000 identities collected from internet, but the videos are mostly similar. So, we choose a subset of 90k videos from about 4000 identities. The images in VoxCeleb2 vary widely in pose but lack variety in expressions and illumination, so we add a custom dataset (ExpressiveFaces) to our training, which contains 3600 videos of 3600 identities. The videos are captured by the users using a hand-held camera (typically the front camera of a mobile phone) as they perform a wide variety of expressions and poses in both indoor and outdoor environments. We sample the videos at 10fps to avoid multiple duplicate frames, randomly delete frames with neutral expression and pose based on a threshold on the expression and pose parameters predicted by \cite{MultiFaceRetarget19}, and then crop the face and extract ground truth 2D landmarks using \cite{MultiFaceRetarget19}. The cropped faces are resized to $224 \times 224$ and grouped into mini-batches, each of $N$ images chosen randomly from different parts of a video to ensure sufficient diversity in the training data. We set $N = 4$ during training and $N=1$ during testing (unless otherwise mentioned) as evaluated to work best for real-time performance in \cite{fml}.

\begin{figure}[t]
    \centering
    \begin{minipage}{0.45\textwidth}
    \centering
    \includegraphics[width=1.0\textwidth, height=0.3\textheight]{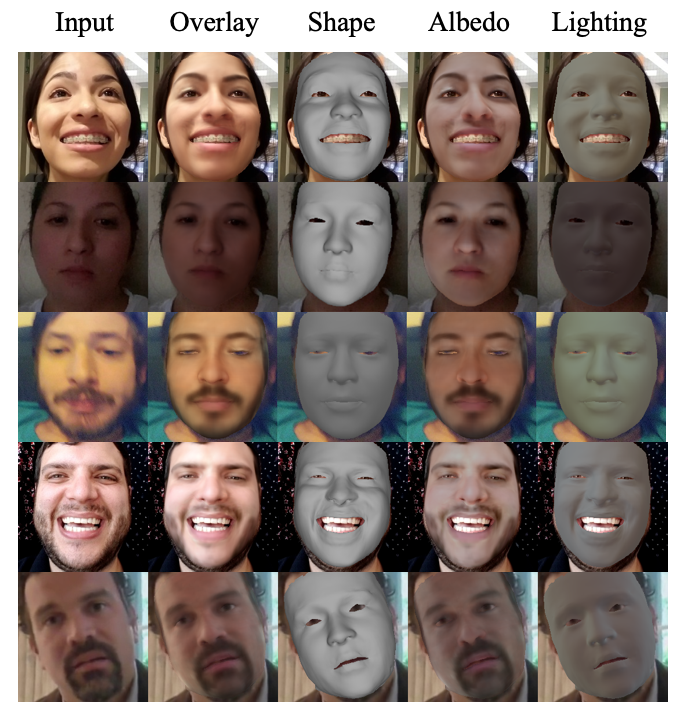}
    \end{minipage}
    \begin{minipage}{0.45\textwidth}
    \centering
    \includegraphics[width=1.0\textwidth,height=0.3\textheight]{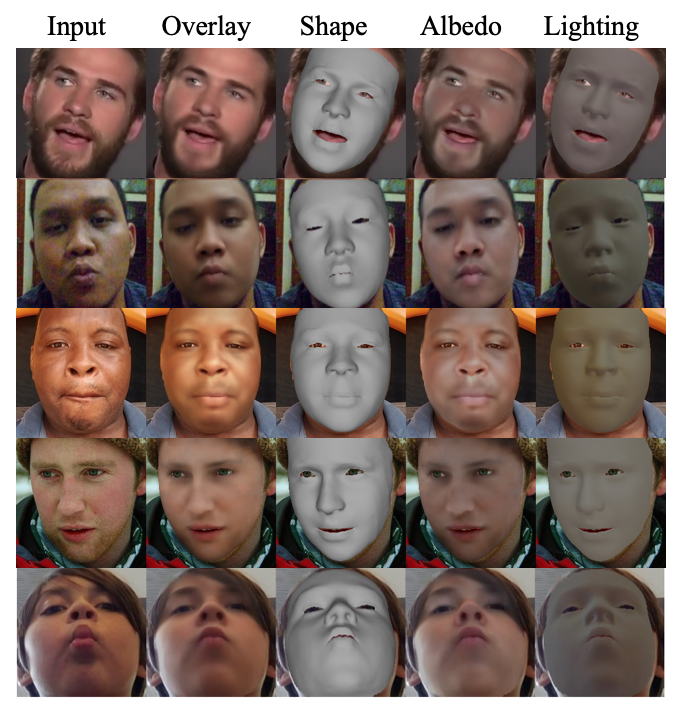}
    \end{minipage}
    \caption[Caption for LOF]{\textbf{Qualitative results of our method}. Our modeling network accurately captures high-fidelity facial details specific to the user, thereby enabling the tracking network to learn user-independent facial motion. Our network can handle a wide variety of pose, expression, lighting conditions, facial hair and makeup etc. Refer to supplementary material for more results for images and videos.}
    \label{fig:results}
\end{figure}

\noindent \textbf{Implementation Details:}
We implemented our networks in Tensorflow and used TF mesh renderer \cite{tfmeshrenderer} for differentiable rendering. During the first stage of training, we train both \textit{TrackNet} and \textit{ModelNet} in an end-to-end manner using equation \ref{eq:loss}. During the second stage of training, we fix the weights of \textit{TrackNet} and fine-tune the \textit{ModelNet} to better learn the expression-specific corrections. The fine-tuning is done using the same loss function as before except the tracking parameter regularization loss since the tracking parameters are now fixed. This training strategy enables us to tackle the bilinear optimization problem of optimizing the blendshapes and the corresponding coefficients, which is generally solved through alternate minimization by existing optimization-based methods. For training, we use a batch size of 8, learning rates of $10^{-4}$ ($10^{-5}$ during second stage) and Adam optimizer for loss minimization. Training takes $\sim$20 hours on a single Nvidia Titan X for the first stage, and another $\sim$5 hours for the second stage. The encoder and decoder together in \textit{ModelNet} has an architecture similar to U-Net \cite{unet} and the encoder in \textit{TrackNet} has the same architecture as ResNet-18 (details in the supplementary). Since our template mesh contains 12264 vertices, we use a corresponding UV map of dimensions $128 \times 128$.

\section{Results}
\begin{figure}[t]
    \centering
    \includegraphics[width=0.95\textwidth]{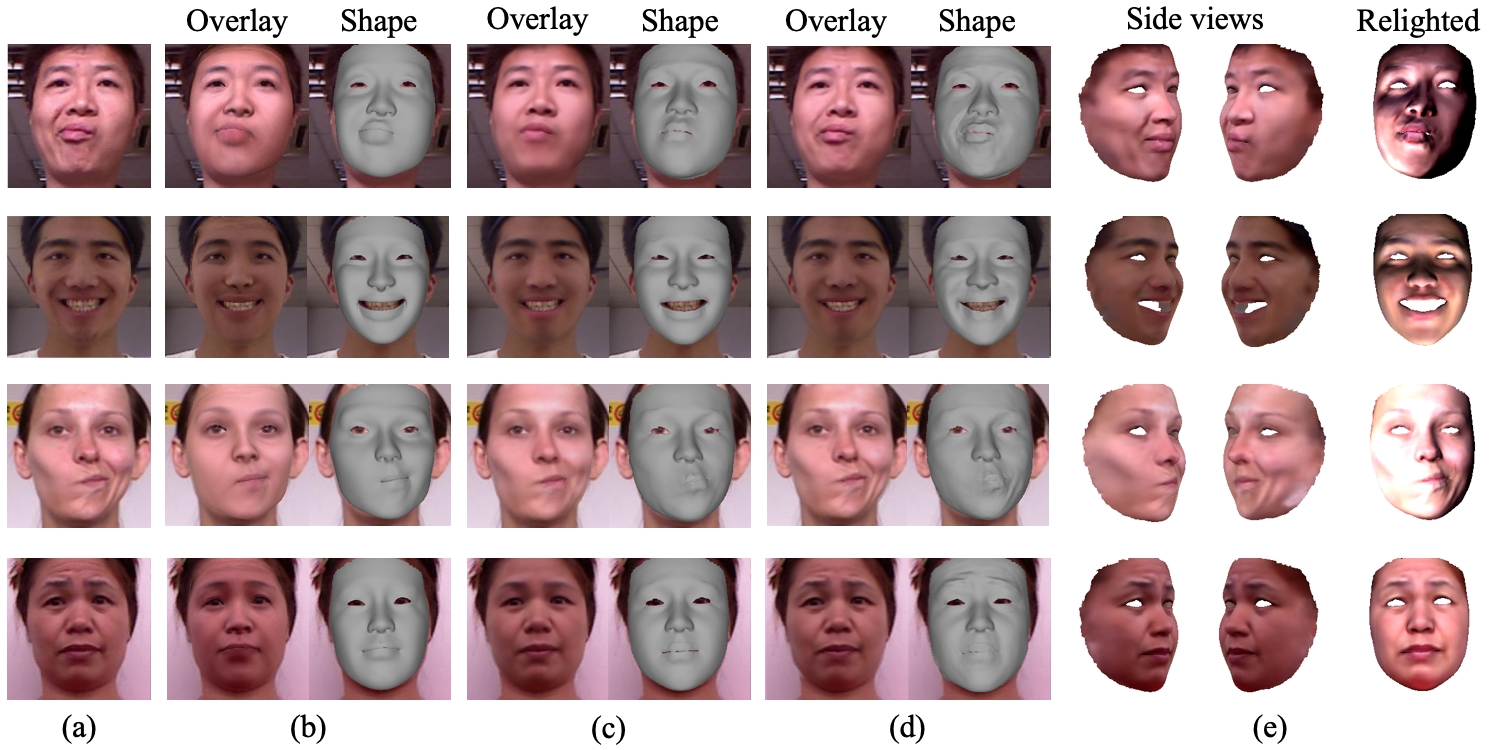}
    \caption{\textbf{Importance of personalization}. (a) input image, (b) reconstruction using 3DMM prior only, (c) reconstruction after adding only identity-based corrections, i.e. $\Delta_0^S$ and $\Delta_0^R$ in eq. (2) and (3) respectively, (d) reconstruction after adding expression-specific corrections, (e) results of (d) with different viewpoints and illumination.}
    \label{fig:personalization}
\end{figure}

We evaluate the effectiveness of our framework using both qualitative results and quantitative comparisons. Fig. \ref{fig:results} shows the personalized face shape and albedo, scene illumination and the final reconstructed 3D face generated from monocular images by our method. Learning a common face shape and albedo from multiple images of a person separately from the image-specific facial motion helps in successfully decoupling the tracking parameters from the learned face model. As a result, our tracking network have the capacity to represent a wide range of expressions, head pose and lighting conditions. Moreover, learning a unified model from multiple images help to overcome issues like partial occlusion, self-occlusion, blur in one or more images. Fig. \ref{fig:personalization} shows a gallery of examples that demonstrate the effectiveness of personalized face modeling for better reconstruction. Fig. \ref{fig:retargetingNerrormaps}a shows that our network can be efficiently used to perform facial motion retargeting to another user or to an external 3D model of a stylized character in addition to face reconstruction.
\subsection{Importance of Personalized Face Model}
\begin{figure}[t]
    \centering
    \includegraphics[width=0.9\textwidth]{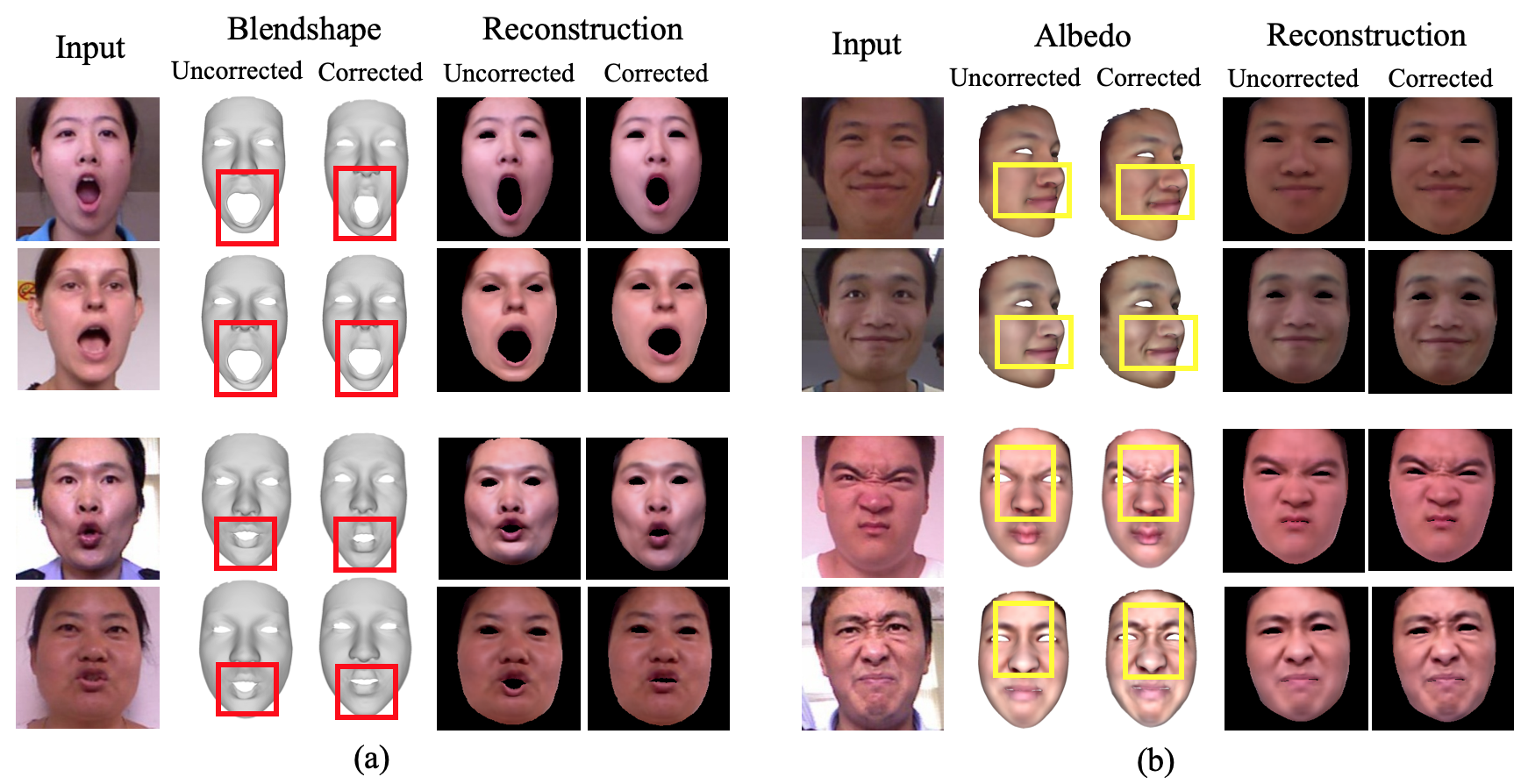}
    \caption{\textbf{Visualization of corrected blendshapes and albedo}. The corrections are highlighted. (a) Learning user-specific blendshapes corrects the mouth shape of the blendshapes, (b) Learning user-specific dynamic albedo maps captures the high-frequency details like skin folds and wrinkles.}
    \label{fig:correction}
\end{figure}

\noindent \textbf{Importance of personalized blendshapes}: Modeling the user-specific local geometry deformations while performing expressions enable our modeling network to accurately fit the facial shape of the input image. Fig. \ref{fig:correction}a shows examples of how the same expression can look different on different identities and how the corrected blendshapes capture those differences for more accurate reconstruction than with the template blendshapes. In the first example, the extent of opening of the mouth in the \textit{mouth open} blendshape is adjusted according to the user expression. In the second example, the mouth shape of the \textit{mouth funnel} blendshape is corrected.

\noindent \textbf{Importance of dynamic textures}: Modeling the user-specific local variations in skin reflectance while performing expressions enable our modeling network to generate a photorealistic texture for the input image. Fig. \ref{fig:correction}b shows examples of how personalized dynamic albedo maps help in capturing the high-frequency expression-specific details compared to static albedo maps. In the first example, our method accurately captures the folds around the nose and mouth during smile expression. In the second example, the unique wrinkle patterns between the eyebrows of the two users are correctly modeled during a disgust expression.
\subsection{Importance of Novel Training Constraints}
\begin{figure}[t]
    \centering
    \includegraphics[width=0.95\textwidth]{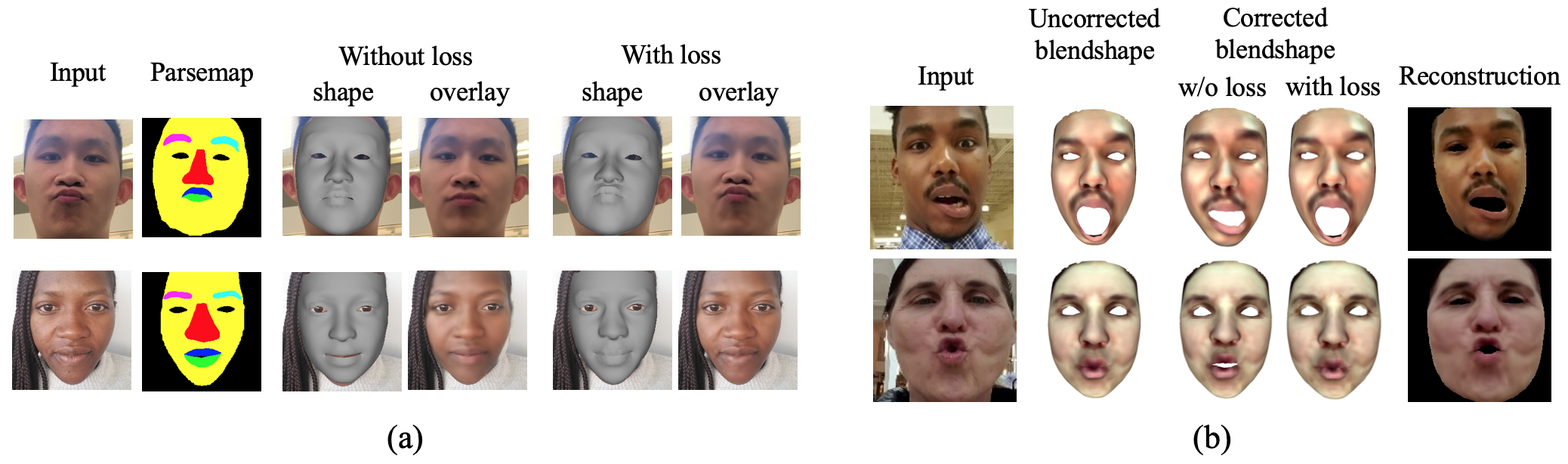}
    \caption{\textbf{Importance of novel training constraints}. (a) importance of face parsing loss in obtaining accurate geometry decoupled from albedo, (b) importance of blendshape gradient loss in retaining the semantic meaning of \textit{mouth open} (row 1) and \textit{kiss} (row 2) blendshapes after correction.}
    \label{fig:parsingNgradient}
\end{figure}

\noindent \textbf{Importance of parsing loss}:
The face parse map ensures that each face part of the reconstructed geometry is accurate as shown in \cite{Zhu_2020_CVPR}. This prevents the albedo to compensate for incorrect geometry, thereby disentangling the albedo from the geometry. However, the authors of \cite{Zhu_2020_CVPR} use parse map in a geometry fitting framework which, unlike our learning framework, does not generalize well to in-the-wild images. Besides, due to the dense correspondence of parse map compared to the sparse 2D landmarks, parsing loss (a) provides a stronger supervision on the geometry, and (b) is more robust to outliers. We demonstrate the effectiveness of face parsing loss in Fig. \ref{fig:parsingNgradient}a. In the first example, the kiss expression is correctly reconstructed with the loss, since the 2D landmarks are not enough to overcome the depth ambiguity. In the second example, without parsing loss the albedo tries to compensate for the incorrect geometry by including the background in the texture. With loss, the nose shape, face contour and the lips are corrected in the geometry, resulting in better reconstruction.

\noindent \textbf{Importance of blendshape gradient loss}:
Even after applying attention masks to restrict the blendshape corrections to local regions, our method can distort a blendshape such that it loses its semantic meaning, which is undesirable for retargeting purposes. We prevent this by enforcing the blendshape gradient loss, as shown in Fig. \ref{fig:parsingNgradient}b. In the first example, without gradient loss, the \textit{mouth open} blendshape gets combined with \textit{jaw left} blendshape after correction in order to minimize the reconstruction loss. With gradient loss, the reconstruction is same but the \textit{mouth open} blendshape retains its semantics after correction. Similarly in the second example, without gradient loss, the \textit{kiss} blendshape gets combined with the \textit{mouth funnel} blendshape, which is prevented by the loss.

\begin{figure}[t]
    \centering
    \includegraphics[width=0.875\textwidth]{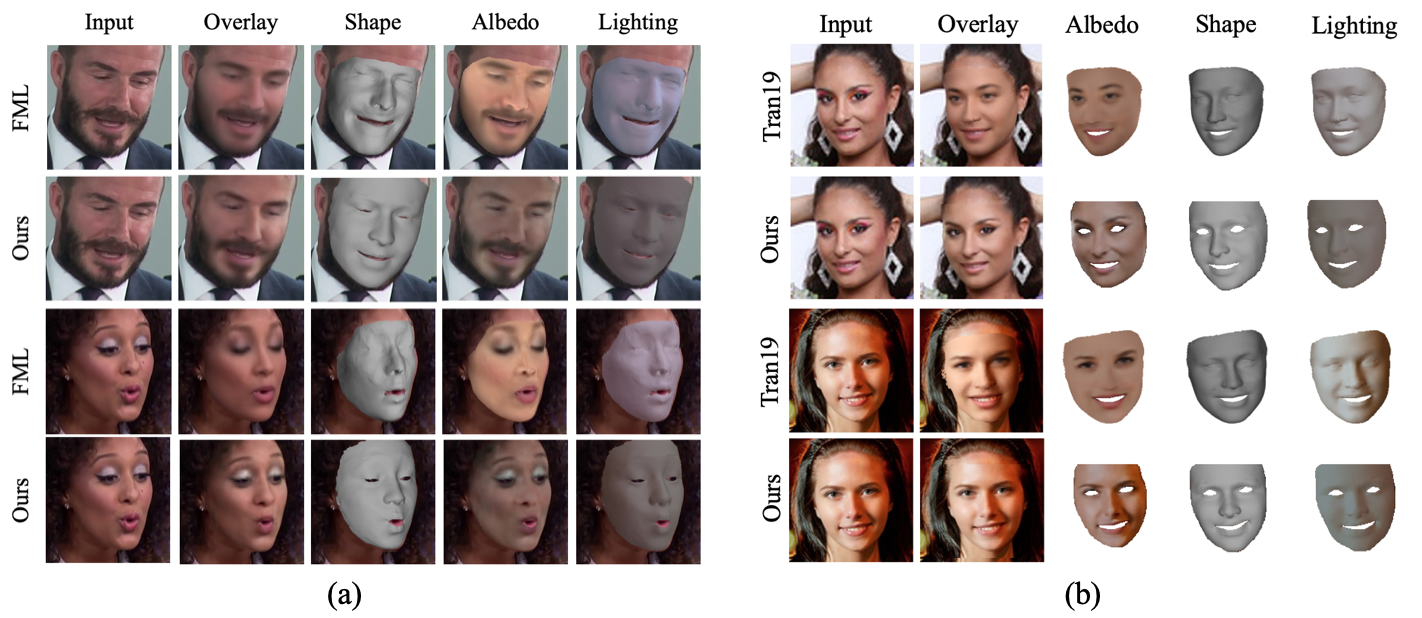}
    \caption{Visual comparison with (a) FML \cite{fml}, (b) Non-linear 3DMM \cite{nonlinear3dmm2019}.}
    \label{fig:sotacompare}
\end{figure}
\subsection{Visual Comparison with State-of-the-art Methods}
\noindent \textbf{3D face reconstruction}:
We test the effectiveness of our method on VoxCeleb2 test set to compare with FML results \cite{fml} as shown in Fig. \ref{fig:sotacompare}a. In the first example, our method captures the mouth shape and face texture better. The second example shows that our personalized face modeling can efficiently model complex expressions like kissing and complex texture like eye shadow better than FML. We also show the visual comparisons between our method and Non-linear 3DMM \cite{nonlinear3dmm2019} on the AFLW2000-3D dataset \cite{3ddfa} in Fig. \ref{fig:sotacompare}b. Similar to FML, Non-linear 3DMM fails to accurately capture the subtle facial details. 

\begin{figure}[t]
    \centering
    \includegraphics[width=1.0\textwidth]{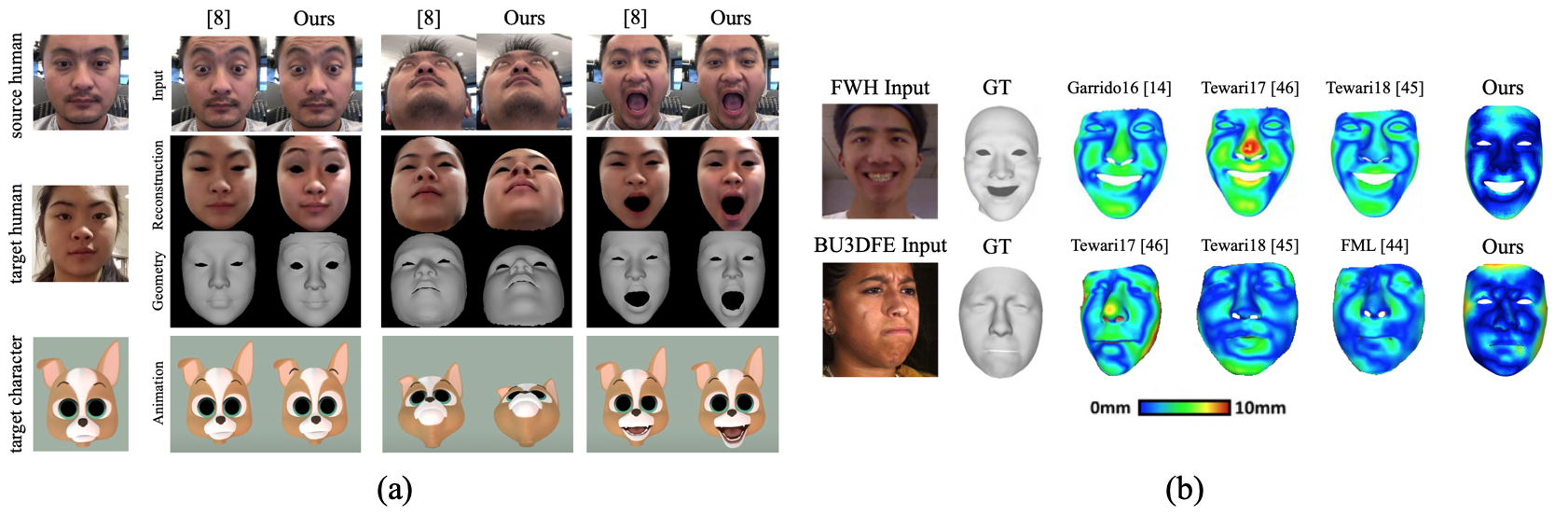}
    \caption{(a) Tracking comparison with \cite{MultiFaceRetarget19}. (b) 3D reconstruction error maps.}
    \label{fig:retargetingNerrormaps}
\end{figure}

\noindent \textbf{Face tracking and retargeting}: By increasing the face modeling capacity and decoupling the model from the facial motion, our method performs superior face tracking and retargeting compared to a recent deep learning based retargeting approach \cite{MultiFaceRetarget19}. Fig. \ref{fig:retargetingNerrormaps}a shows some frames from a user video and how personalization helps in capturing the intensity of the expressions more accurately.

\subsection{Quantitative Evaluation}
\begin{table}[t]
    \caption{\textbf{Ablation study}. Evaluation of different components of our proposed method in terms of standard evaluation metrics. Note that B and C are obtained with all the loss functions other than the parsing loss and the gradient loss.}
    \centering
    \label{tab:ablation}
    \scalebox{0.75}{
    \begin{tabular}{|l|c|c|c|c|c|c|c|}
    \hline
    \multirow{3}{*}{\textbf{Method}} & \multicolumn{4}{|c|}{\textbf{3D error (mm)}} & \textbf{NME} & \textbf{AUC} & \textbf{Photo error} \\ \cline{2-8}
                & \multicolumn{2}{|c|}{\textbf{BU-3DFE}} & \multicolumn{2}{|c|}{\textbf{FWH}} & \textbf{AFLW2000-3D} & \textbf{300VW} & \textbf{CelebA*} \\ \cline{2-8}
                & Mean & SD & Mean & SD & Mean & Mean & Mean\\
    \hline
    3DMM prior (A)      & 2.21 & 0.52 & 2.13 & 0.49 & 3.94 & 0.845 & 22.76 \\ \hline
    A + blendshape corrections (B)   & 2.04 & 0.41 & 1.98 & 0.44  & 3.73 & 0.863 & 22.25  \\ \hline 
    B + albedo corrections (C)     & 1.88 & 0.39 & 1.85 & 0.41  & 3.68 & 0.871 & 20.13 \\ \hline 
    C + parsing loss (D)            & 1.67 & 0.35 & 1.73 & 0.37  & 3.53 & 0.883 & 19.49 \\ \hline 
    D + gradient loss (final)   & 1.61 & 0.32 & 1.68 & 0.35  & 3.49 & 0.890 & 18.91 \\ \hline 
    \end{tabular}}
\centering
\caption{\textbf{Quantitative evaluation with state-of-the-art methods}. (a) 3D reconstruction error (mm) on BU-3DFE and FWH datasets, (b) NME (\%) on AFLW2000-3D (divided into 3 groups based on yaw angles), (c) AUC for cumulative error distribution of the 2D landmark error for 300VW test set (divided into 3 scenarios by the authors). Note that higher AUC is better, whereas lower value is better for the other two metrics.}
    \begin{subtable}{.33\linewidth}
    \centering
        \caption{}
        \label{tab:3derror}
        \scalebox{0.75}{
        \begin{tabular}{|l|c|c|c|c|}
        \hline
        \multirow{2}{*}{\textbf{Method}} & \multicolumn{2}{|c|}{\textbf{BU-3DFE}} & \multicolumn{2}{|c|}{\textbf{FWH}} \\ \cline{2-5}
                        & Mean & SD & Mean & SD\\
        \hline
        \cite{selfsupervised_ayush}  & 1.83 & 0.39 & 1.84 & 0.38 \\ \hline
        \cite{tewari17MoFA}    & 3.22 & 0.77 & 2.19 & 0.54 \\ \hline 
        \cite{fml}    & 1.74 & 0.43 & 1.90 & 0.40 \\ \hline 
        Ours         & \textbf{1.61} & \textbf{0.32} & \textbf{1.68} & \textbf{0.35} \\ \hline 
        \end{tabular}}
    \end{subtable}
    \begin{subtable}{.35\linewidth}
    \caption{}
    \label{table:aflw_nme}
    \centering
        \scalebox{0.75}{
        \begin{tabular}{|l|c|c|c|c|}
        \hline
         \textbf{Method}  &  [0-30\degree] & [30-60\degree] & [60-90\degree] & Mean\\
         \hline
        \cite{3ddfa} & 3.43 & 4.24 & 7.17 & 4.94\\
         \hline
        \cite{FasterTR} & 3.15 & 4.33 & 5.98 & 4.49\\
         \hline
        \cite{prn} & 2.75 & 3.51 & 4.61 & 3.62\\
         \hline
         \cite{MultiFaceRetarget19} & 2.91 & 3.83 & 4.94 & 3.89 \\
         \hline
         Ours & \textbf{2.56} & \textbf{3.39} & \textbf{4.51} & \textbf{3.49}\\
         \hline
        \end{tabular}}
        
    \end{subtable}
    \begin{subtable}{0.3\linewidth}
        \centering
        \caption{}
        \label{table:300vw_auc}
        \scalebox{0.75}{
        \begin{tabular}{|l|c|c|c|}
        \hline
         \textbf{Method} & Sc. 1 & Sc. 2 & Sc. 3\\
         \hline
        \cite{yangetal} & 0.791 & 0.788 & 0.710\\
          \hline
        \cite{MTCNN} & 0.748 & 0.760 & 0.726\\
          \hline
        \cite{Jointalignment} & 0.847 & 0.838 & 0.769\\
          \hline
          \cite{MultiFaceRetarget19} & 0.901 & 0.884 & 0.842\\
          \hline
          Ours & \textbf{0.913} & \textbf{0.897} & \textbf{0.861}\\
          \hline
        \end{tabular}}
        
    \end{subtable}
    \centering
    \caption{Quantitative evaluation of retargeting accuracy (measured by the expression metric) on \cite{MultiFaceRetarget19} expression test set. Lower error means the model performs better for extreme expressions.}
    \label{table:exp}
    \scalebox{0.8}{
    \begin{tabular}{|l|l|l|l|l|l|l|l|l|l|l|}
    \hline
    \multicolumn{1}{|c|}{Model} & \multicolumn{1}{c|}{\begin{tabular}[c]{@{}c@{}}Eye\\ Close\end{tabular}} & \multicolumn{1}{c|}{\begin{tabular}[c]{@{}c@{}}Eye\\ Wide\end{tabular}} & \multicolumn{1}{c|}{\begin{tabular}[c]{@{}c@{}}Brow\\ Raise\end{tabular}} & \multicolumn{1}{c|}{\begin{tabular}[c]{@{}c@{}}Brow\\ Anger\end{tabular}} & \multicolumn{1}{c|}{\begin{tabular}[c]{@{}c@{}}Mouth\\ Open\end{tabular}} & \multicolumn{1}{c|}{\begin{tabular}[c]{@{}c@{}}Jaw\\ L/R\end{tabular}} & \multicolumn{1}{c|}{\begin{tabular}[c]{@{}c@{}}Lip\\ Roll\end{tabular}} & Smile & Kiss & \multicolumn{1}{c|}{Avg} \\ \hline
    (1) Retargeting \cite{MultiFaceRetarget19}  & 0.117  & 0.407   & 0.284   & 0.405   & 0.284   & 0.173   & 0.325   & 0.248   & 0.349 & 0.288 \\ \hline
    (2) Ours          & 0.140   & 0.389   & 0.259   & 0.284   & 0.208   & 0.394   & 0.318   & 0.121   & 0.303 & \textbf{0.268} \\ \hline
    \end{tabular}
    }
\end{table}

\noindent \textbf{3D face reconstruction}: We compute 3D geometry reconstruction error (root mean squared error between a predicted vertex and ground truth correspondence point) to evaluate our predicted mesh on 3D scans from BU-3DFE \cite{yin20063d} and Facewarehouse (FWH) \cite{Cao3d}. For BU-3DFE we use both the views per scan as input, and for FWH we do not use any special template to start with, unlike Asian face template used by FML. Our personalized face modeling and novel constraints together result in lower reconstruction error compared to state-of-the-art methods (Table \ref{tab:3derror} and Fig. \ref{fig:retargetingNerrormaps}b). The optimization-based method \cite{3Dfacerig16} obtains 1.59mm 3D error for FWH compared to our 1.68mm, but is much slower (120s/frames) compared to our method (15.4ms/frame). We also show how each component of our method helps in improving the overall output in Table \ref{tab:ablation}. For photometric error, we used 1000 images of CelebA \cite{celebA} (referred as CelebA*) dataset to be consistent with \cite{fml}.

\noindent \textbf{Face tracking}: We evaluate the tracking performance of our method using two metrics: 1) Normalized Mean Error (NME), defined as an average Euclidean distance between the 68 predicted and ground truth 2D landmarks normalized by the bounding box dimension, on AFLW2000-3D dataset \cite{3ddfa}, and 2) Area under the Curve (AUC) of the cumulative error distribution curve for 2D landmark error \cite{Jointalignment} on 300VW video test set \cite{300VW}. Table \ref{table:aflw_nme} shows that we achieve lower landmark error compared to state-of-the-art methods although our landmarks are generated by a third-party method. We also outperform existing methods on video data (Table \ref{table:300vw_auc}). For video tracking, we detect the face in the first frame and use the bounding box from previous frame for subsequent frames similar to \cite{MultiFaceRetarget19}. However, the reconstruction is performed on a per-frame basis to avoid inconsistency due to the choice of random frames.

\noindent \textbf{Facial motion retargeting}: In order to evaluate whether our tracked facial expression gets correctly retargeted on the target model, we use the expression metric defined as the mean absolute error between the predicted and ground truth blendshape coefficients as in \cite{MultiFaceRetarget19}. Our evaluation results in Table \ref{table:exp} emphasize the importance of personalized face model in improved retargeting, since \cite{MultiFaceRetarget19} uses a generic 3DMM as its face model.

\section{Conclusion}

We propose a novel deep learning based approach that learns a user-specific face model (expression blendshapes and dynamic albedo maps) and user-independent facial motion disentangled from each other by leveraging in-the-wild videos. Extensive evaluation have demonstrated that our personalized face modeling combined with our novel constraints effectively performs high-fidelity 3D face reconstruction, facial motion tracking, and retargeting of the tracked facial motion from one identity to another.

\textbf{Acknowledgements:} We thank the anonymous reviewers for their constructive feedback, Muscle Wu, Wenbin Zhu and Zeyu Chen for helping, and Alex Colburn for valuable discussions.

\bibliographystyle{splncs04}

\appendix

\section{Appendix}

Our two main contributions in this paper are a) to model personalized expression blendshapes and dynamic albedo maps, and b) to perform joint tracking and modeling in a decoupled manner to support both reconstruction and motion retargeting. In this Appendix, we provide more qualitative results of our approach to substantiate the improvements over baseline caused by our contributions. 

\subsection{Training Details}
The architectures of the encoder and decoders of our modeling network are given in Table \ref{tab:encoderarchitecture} and Table \ref{tab:decoderarchitecture}. Each Conv2D and Deconv2D layer is followed by batch normalization which is then followed by ReLU activation. Our end-to-end network has a size of 240 MB and takes 15.4 ms to execute 1 image and 37.5 ms to execute 4 images on a Titan X GPU on average. The loss weights are chosen to be: $\lambda_{\text{ph}} = 200; \lambda_{\text{lm}} = 0.1; \lambda_{\text{pa}} = 50; \lambda_{\text{sd}} = 2.5; \lambda_{\text{bg}} = 1.5; \lambda_{\text{reg}} = 10^{-3}; \lambda_\gamma = 0.02$.

Fine-tuning the modeling network during the second stage of training ensures further decoupling between tracking and modeling. Besides, our method of obtaining the user-specific face shape and albedo from multiple frames helps in learning the static shape and albedo corrections separately from the expression-specific shape and albedo variations. As a result, our framework can produce photorealistic expression-specific deformations on a new user during testing.

\begin{figure*}
\centering
\includegraphics[width=0.6\textwidth]{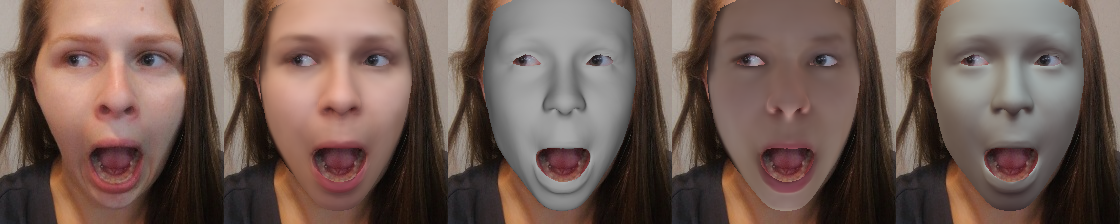}\\
\includegraphics[width=0.6\textwidth]{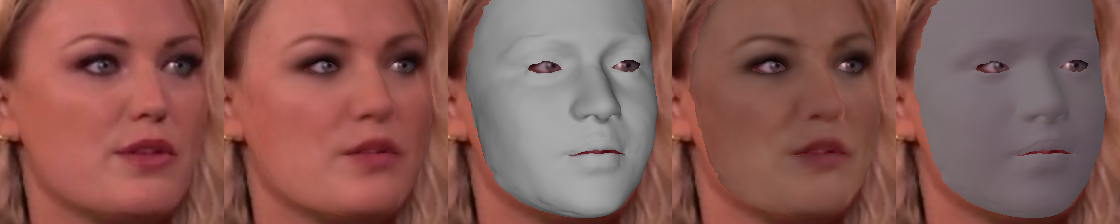}\\
\includegraphics[width=0.6\textwidth]{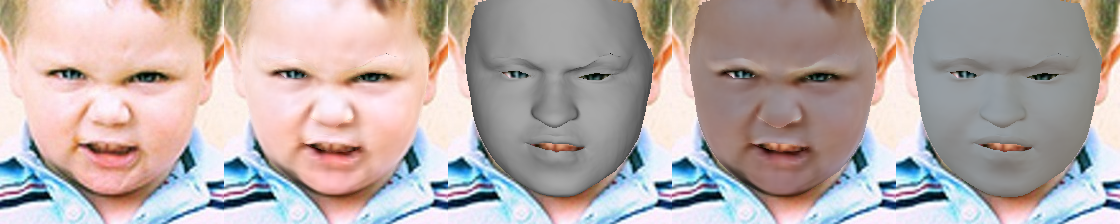}\\
\includegraphics[width=0.6\textwidth]{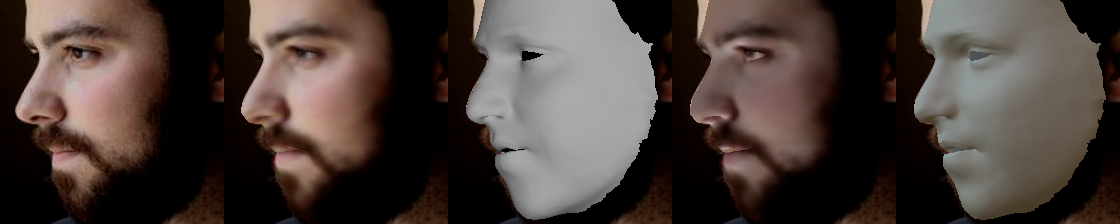}\\
\includegraphics[width=0.6\textwidth]{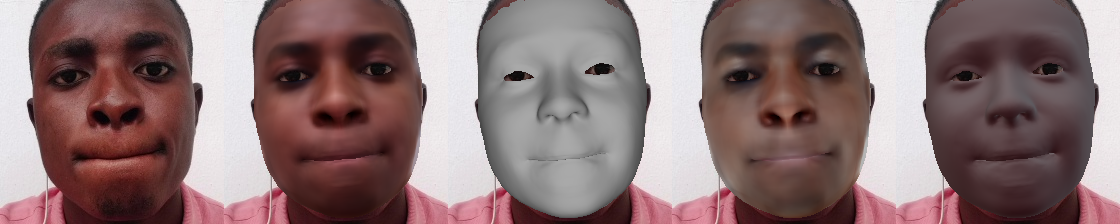}\\
\includegraphics[width=0.6\textwidth]{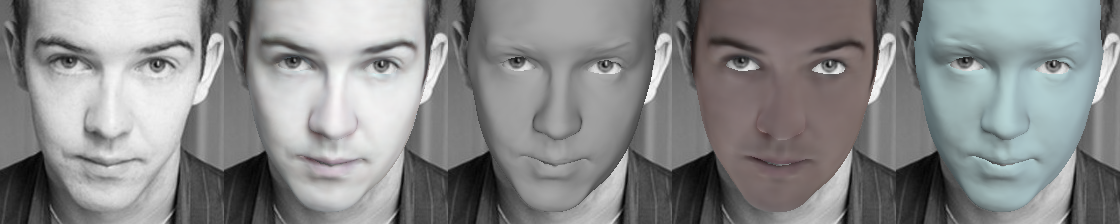}\\
\includegraphics[width=0.6\textwidth]{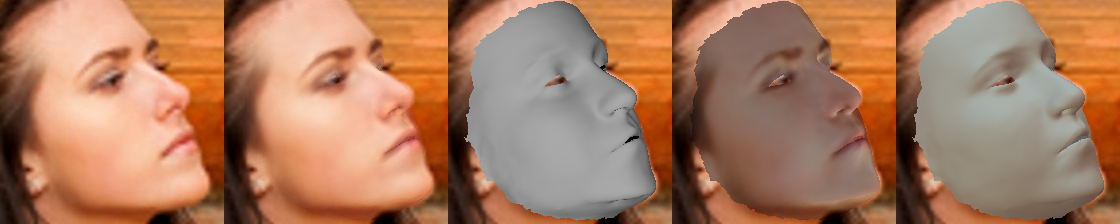}\\
\includegraphics[width=0.6\textwidth]{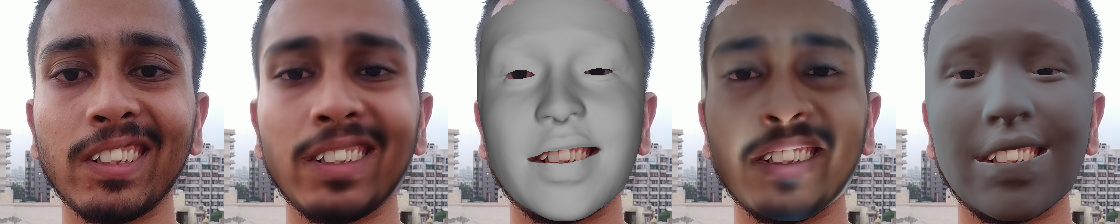}\\
\includegraphics[width=0.6\textwidth]{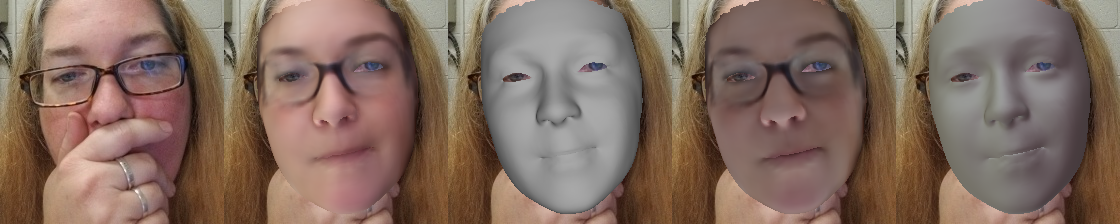}\\
\includegraphics[width=0.6\textwidth]{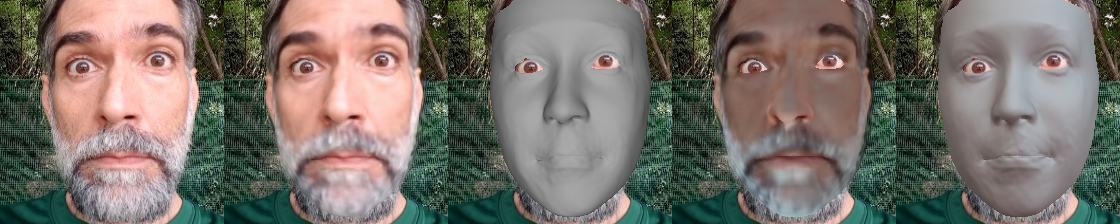}\\
\includegraphics[width=0.6\textwidth]{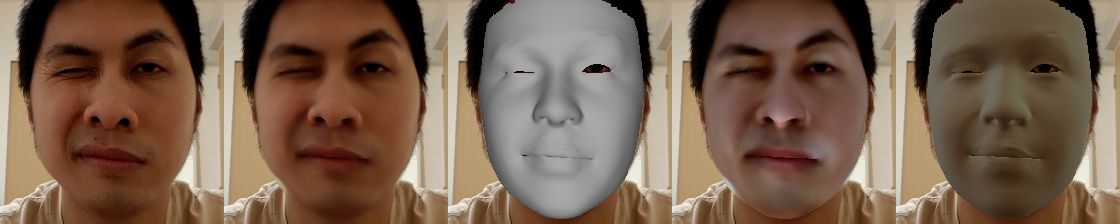}\\
\includegraphics[width=0.6\textwidth]{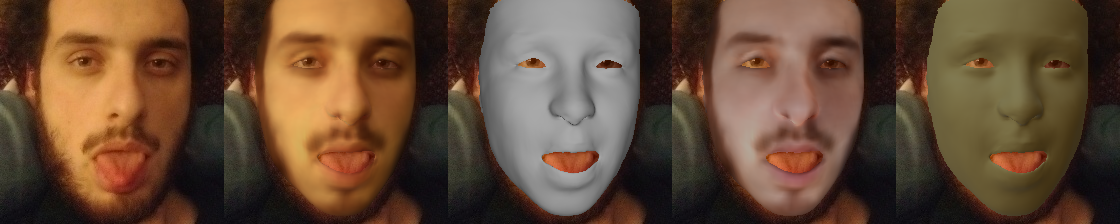}
\caption{Face reconstruction results using our method on our test set. From left to right: input image, overlay, shape, albedo, lighting.}
\label{fig:ourvideos}
\end{figure*}

\subsection{More Qualitative Results}
Fig. \ref{fig:ourvideos} shows 3D face reconstruction results using our method on our test data. It can be noted that our method can reconstruct faces accurately even under conditions like unusual lighting (row 6), uncommon face shapes (baby face in row 3), extreme poses (rows 4 and 7), occlusion (row 9), extreme expressions etc. However, similar to \cite{fml}, our method embeds eye glasses into the albedo (row 9). We would like to point out that videos of ExpressiveFaces are captured by hand-held cameras and have resolution of $1920 \times 1080$, from which we crop faces resized to size $224 \times 224$. On the other hand, videos in Voxceleb2 \cite{voxceleb2} are scraped from Youtube and hence have a very different distribution (lower resolution) than the videos of ExpressiveFaces.

\begin{table}[t]
    \caption{\textbf{Architecture of our networks}. $s\#$ refers to stride $\#$. (a) Architecture of the shared encoder of our modeling network. The outputs of the encoder for each input image in a mini-batch are average pooled to obtain a single $(7,7,512)$ feature that becomes the input to both the decoders. (b) Architecture of each of the decoder of our modeling network. Note that the output of the last Deconv2D layer goes into both the last 2 Conv2D layers.}
    \centering
    \begin{subtable}{0.49\linewidth}
        \centering
        \caption{}
        \label{tab:encoderarchitecture}
        \scalebox{0.75}{\begin{tabular}{|c|c|c|}
        \hline
        \textbf{Layers} & \textbf{Input Shape} & \textbf{Output Shape} \\ \hline
        Conv2D ($7 \times 7$, s2) & (224,224,3) & (112,112,64)\\ \hline
        Maxpool ($3 \times 3$, s2) & (112,112,64) & (56,56,64)\\ \hline
        Conv2D ($3 \times 3$, s1) & (56,56,64) & (56,56,128)\\ \hline
        Conv2D ($3 \times 3$, s2) & (56,56,128) & (28,28,128)\\ \hline
        Conv2D ($3 \times 3$, s1) & (28,28,128) & (28,28,256)\\ \hline
        Conv2D ($3 \times 3$, s2) & (28,28,256) & (14,14,256)\\ \hline
        Conv2D ($3 \times 3$, s1) & (14,14,256) & (14,14,512)\\ \hline
        Conv2D ($3 \times 3$, s2) & (14,14,512) & (7,7,512)\\ \hline
        \end{tabular}}
    \end{subtable}
    \begin{subtable}{0.49\linewidth}
        \centering
        \caption{}
        \label{tab:decoderarchitecture}
        \scalebox{0.75}{\begin{tabular}{|c|c|c|}
        \hline
        \textbf{Layers} & \textbf{Input Shape} & \textbf{Output Shape}\\ \hline
        Deconv2D ($4 \times 4$, s2) & (7,7,512) & (16,16,512)\\ \hline
        Deconv2D ($4 \times 4$, s2) & (16,16,512) & (32,32,256)\\ \hline
        Deconv2D ($4 \times 4$, s2) & (32,32,256) & (64,64,128)\\ \hline
        Deconv2D ($4 \times 4$, s2) & (64,64,128) & (128,128,64)\\ \hline
        Conv2D ($1 \times 1$, s1) & (128,128,64) & (128,128,3)\\ \hline
        Conv2D ($1 \times 1$, s1) & (128,128,64) & (128,128,56*3) \\ \hline
        \end{tabular}}
    \end{subtable}
    
\end{table}

\subsection{Video Results}
The performance of our method on face videos is shown in the video attached with this document. We show three applications of our method: a) personalized reconstruction, b) retargeting to a different user's face model, and c) retargeting to an external 3D puppet. To process the input video, we detect the face bounding box using \cite{MultiFaceRetarget19} for the first frame only. For the subsequent frames, the bounding box of each frame is obtained from the boundaries of the 2D landmarks predicted in the previous frame. This technique helps in reducing the jitter in the results due to inconsistent bounding box selection if done on a per-frame basis. However, some temporal smoothing as a post-processing step will produce better results.

\end{document}